# Assessment of effective parameters on dilution using approximate reasoning methods in longwall mining method, Iran coal mines

H.Owladeghaffari, K.Shahriar & GH.R.Saeedi
*Department of Mining &Metallurgical Engineering, Amirkabir university of Technology, Tehran, Iran*

ABSTRACT: Approximately more than 90% of all coal production in Iranian underground mines is derived directly longwall mining method. Out – of – seam dilution (namely 20-30 %) is one of the essential problems in these mines. Therefore the dilution can impose the additional cost of mining and milling. As a result, recognition of the effective parameters on the dilution has a remarkable role in industry. In this way, this paper has analyzed the influence of 13 parameters (attributed variables) versus the decision attribute (dilution value), so that using two approximate reasoning methods, namely Rough Set Theory (RST) and Self Organizing Neuro-Fuzzy Inference System (SONFIS) the best rules on our collected data sets has been extracted. The other benefit of later methods is to predict new unknown cases. So, the reduced sets (reducts) by RST have been obtained. Therefore the emerged results by utilizing mentioned methods shows that the high sensitive variables are "thickness of layer, length of stope, rate of advance, number of miners, type of advancing".

## 1 INTRODUCTION

The dilution, as employed in mining, is on the reduction of the content of applicable constituents in the extracted ore as compared to their proportion in the mass of ore in the place (Popov, 1971). The main route of dilution can be ensued in extraction step where imminent extra costs in several levels of mining, mineral processing and environmental will be inevitable. As well as other applied methods in different streams of mining engineering to analysis of rock based system, in investigation of dilution, so, two known strategies cab be highlighted: direct methods and indirect methods. In first situation one can consider the real and simplified parameters on the dilution in order to the real mining procedure system are transferred in to the accessible mathematical model. The distinguished approaches, in this field, are analytical and numerical modeling (Henning&Mitri, 2000) while former option is based on the direct calculations and latter case come from "computational" processes. In facing of theses methodologies, indirect emulation of an event is based upon the humanity performance as well as historical aggregation of experts' experiences or the mimicked intelligent systems upon the individuality or society behaviors of glory creatures. Due to being of many agents in the recognizing of most main parameters on the dilution, representing of several definitions and relations ,along successive years, has been unavoidable for example see (Pakalnis et al, 1995).

Considering of the thumb relation using simple and available parameters can be presumed as an overall measure that covers direct and indirect operators on the decision part. In fact, such a relation is an escape road from the complexity of problem while other unapparent factors take marginal guise. To bate of this fact, we refer to the reactions of real natural complex systems with their intricate circumstances. The emergence of such systems can be appeared in natural computing methods (not 1-1 mapping models) involved methods inspired in our nature as well as in microscopic or macroscopic view, i.e., neural computations, approximate reasoning(AR) methods(handling of vagueness and uncertainty,…), evolutionary algorithms, artificial life, and artificial immune systems, quantum computing and so forth. In this study we employ approximate reasoning methods to analysis of dilution and based upon two main subsets of AR, Rough Set Theory and Fuzzy inference system, the position of effective parameters are recognizes and evaluated.

## 2 PERSPECTIVE OF METHODS

To dissection of different attributes effects on the dilution, we'll use some of the main approaches in the data engineering field, as well associated with intelligent computational and approximate reasoning methods which based on them a new algorithm SONFIS will be proposed.

### 2.1 Rough set theory (RST)

The rough set theory introduced by Pawlak (Pawlak1981, 1982) has often proved to be an excellent mathematical tool for the analysis of a vague description of object. The adjective vague referring to the quality of information means inconsistency, or ambiguity which follows from information granulation. An information system is a pair $S=< U, A >$, where $U$ is a nonempty finite set called the universe and $A$ is a nonempty finite set of attributes. An attribute a can be regarded as a function from the domain $U$ to some value set $V_a$. An information system can be represented as an attribute-value table, in which rows are labeled by objects of the universe and columns by attributes. With every subset of attributes $B \subseteq A$, one can easily associate an equivalence relation $I_B$ on $U$:

$$I_B = \{(x,y) \in U : for\ every\ a \in B, a(x) = a(y)\} \qquad (1)$$

Then, $I_B = \bigcap_{a \in B} I_a$.

If $X \subseteq U$, the sets $\{x \in U : [x]_B \subseteq X\}$ and $\{x \in U : [x]_B \cap X \neq \varphi\}$, where $[x]_B$ denotes the equivalence class of the object $x \in U$ relative to $I_B$, are called the B-lower and the B-upper approximation of X in S and denoted by $\underline{BX}$ and $\overline{BX}$, respectively. Consider $U = \{x_1, x_2, ..., x_n\}$ and $A = \{a_1, a_2, ..., a_n\}$ in the information system $S = \prec U, A \succ$. By the discernibility matrix $M(S)$ of $S$ is meant an $n*n$ matrix such that

$$c_{ij} = \{a \in A : a(x_i) \neq a(x_j)\} \qquad (2)$$

A discernibilty function $f_s$ is a function of m Boolean variables $a_1...a_m$ corresponding of attribute $a_1...a_m$, respectively, and defined as follows:

$$f_s(a_1,...,a_m) = \wedge\{\vee(c_{ij}) : i, j \leq n, j \prec i, c_{ij} \neq \varphi\} \qquad (3)$$

where $\vee(c_{ij})$ is the disjunction of all variables with $a \in c_{ij}$. Using such discriminant matrix the appropriate rules are elicited (Pal et al, 2004). Indiscernibility relation (similarity) reduces the data by defining of equivalence classes, under the present attributes. Reduction of attributes (conditions) can be utilized by keeping of the attributes that maintain the similarity relation. The underway of idea behind finding out of reducts is the preserving of general information state with minimum attributes, and so superfluous attributes are eliminated. In equation (3), it can be proved that $\{a_{i1},...,a_{ir}\}$ is a reduct in $S$ iff $a_{i1} \wedge ... \wedge a_{ir}$ is a prime implicant of $f_s$. In our work, rule induction procedure was implemented within the ROSETA software system, a rough set tool-kit for knowledge discovery and data mining (φhm et al,1998).

### 2.2 Neuro-Fuzzy Inference System (NFIS)

There are different alternatives of fuzzy inference systems. Two well-known fuzzy modeling methods are the Tsukamoto fuzzy model and Takagi– Sugeno–Kang (TSK) model. In the present study, we focus on the TSK. The TSK fuzzy inference systems can be impl anted in the form of a Neuro-fuzzy network structure. In this study, we have employed an *adaptive neuro-fuzzy inference system* (Jang et al, 1997), within the general structure of SONFIS.

One of the most important stages of the Neuro-fuzzy TSK network generation is the establishment of the inference rules. Often employed method is used the so-called grid method, in which the rules are defined as the combinations of the membership functions for each input variable. If we split the input variable range into a limited number (say $n_i$ for $i=1, 2... n$) of membership functions, the combinations of them lead to many different inference rules.

The problem is that these combinations correspond in many cases to the regions of no data, and hence a lot of them may be deleted. This problem can be solved by using the fuzzy self-organization algorithm. This algorithm splits the data space into a specified number of overlapping clusters. Each cluster may be associated with the specific rule of the center corresponding to the center of the appropriate cluster. In this way all rules correspond to the regions of the space-containing majority of data and the problem of the empty rules can be avoided. The ultimate goal of data clustering is to partition the data into similar subgroups. This is accomplished by employing some similar measures -e.g., the Euclidean distance-(Nauk et al, 1997).

In this paper data clustering is used to derive membership functions from measured data, which, in turn, determine the number of *If-Then rules* in the model (i.e., rules indication).

The method employed in this paper is the subtractive clustering method, proposed by Chiu as one of the simplest clustering methods (Chiu, 1994).

### 2.3 *Self-Organizing feature Map (SOM)*

Kohonen's SOM algorithm has been well renowned as an ideal candidate for classifying input data in an unsupervised learning way. Kohonen self-organizing networks (Kohonen feature maps or topology-preserving maps) are competition-based network paradigm for data clustering. The learning procedure of Kohonen feature maps is similar to the competitive learning networks. The main idea behind competitive learning is simple; the winner takes all.

The competitive transfer function returns neural outputs of 0 for all neurons except for the winner which receives the highest net input with output 1. SOM changes all weight vectors of neurons in the near vicinity of the winner neuron towards the input vector. Due to this property SOM, are used to reduce the dimensionality of complex data (data clustering). Competitive layers will automatically learn to classify input vectors, the classes that the competitive layer finds are depend only on the distances between input vectors (Kohonen, 1986).

### 2.4 *A combining of SOM&NFIS: SONFIS*

In this part, we reproduce the proposed a hybrid intelligent algorithm in (Owladeghaffari et al, 2008):

Step (1): dividing the monitored data into groups of training and testing data
Step (2): first granulation (clustering) by SOM or other crisp granulation methods
    Step (2-1): selecting the level of granularity randomly or depend on the obtained error from the NFIS or RST (regular neuron growth)
    Step (2-2): construction of the granules (no-fuzzy clusters).
Step (3): second granulation by NFIS or RST
    Step (3-1): crisp granules as a new data.
    Step (3-2): selecting the level of granularity; (Error level, number of rules, strength threshold...)
    Step (3-3): checking the suitability. (Close-open iteration: referring to the real data and reinspect closed world)
    Step (3-4): construction of fuzzy/rough granules.
Step (4): extraction of knowledge rules

Balancing assumption is satisfied by the close-open iterations: this process is a guideline to balancing of crisp and sub fuzzy/rough granules by some random/regular selection of initial granules or other optimal structures and increment of supporting rules (fuzzy partitions or increasing of lower /upper approximations ), gradually. In this study, we use only two known intelligent systems: SOM&NFIS. The overall schematic of Self Organizing Neuro-Fuzzy Inference System -Random -: SONFIS-R has been shown in figure1.

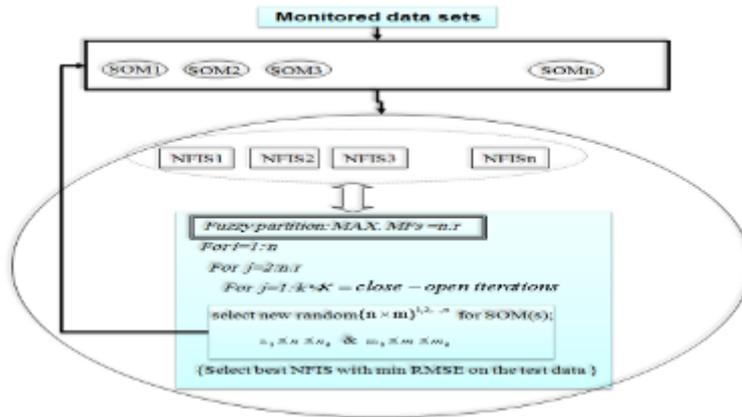

Figure 1. Self Organizing Neuro-Fuzzy Inference System: SONFIS-R (Owladeghaffari et al, 2008)

3 RESULTS

Figure 2 shows an overall view of the collected data set in a few coal mines, Iran.
To employing of the mentioned methods, we must ascribe some appropriate codes to some qualitative attributes (table 1). In first situation, to obtain most main conditional parameters on the dilution, we use RST. After this step, by applying SONFIS-R the simplest rules companion with their memberships functions are displaced. So, on those rules graphical relations of reduct's members are highlighted. In the whole of this paper, we select 21 objects among 30 patterns as training data set and the remained objects get testing data facet.
With considering this point that the creation of discernible matrix-in RST- is depend on the transferring of data in to the arbitrary-or best- ranges (bins)-symbolic values-, we employ one dimensional topology grid SOM, in which attributes are transferred within 3 categories: *low (1), medium (2) and high (3)* (fig3).Then, the categorized attributes as an information table are assessed by the RST.

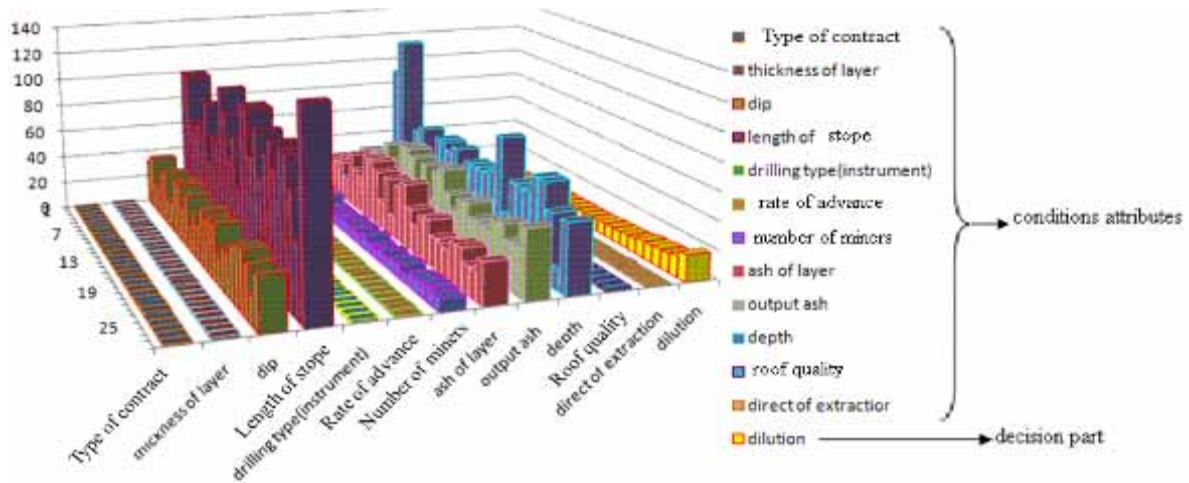

Figure 2.3D Column view of the accumulated data set and the appropriate decision table

Table1.attributed codes to some parameters

| Attributes | Ascribed codes |
|---|---|
| Contract Type | Contract work=1 ;Service(state)=2 |
| Drilling Instrument | Pic=1 ;Drilling &blasting=2 |
| Direct of Extraction | Forward=1 ;Backward=2 |
| Type of Floor Rock | Argillite=1 ;Sandy rock=2 |

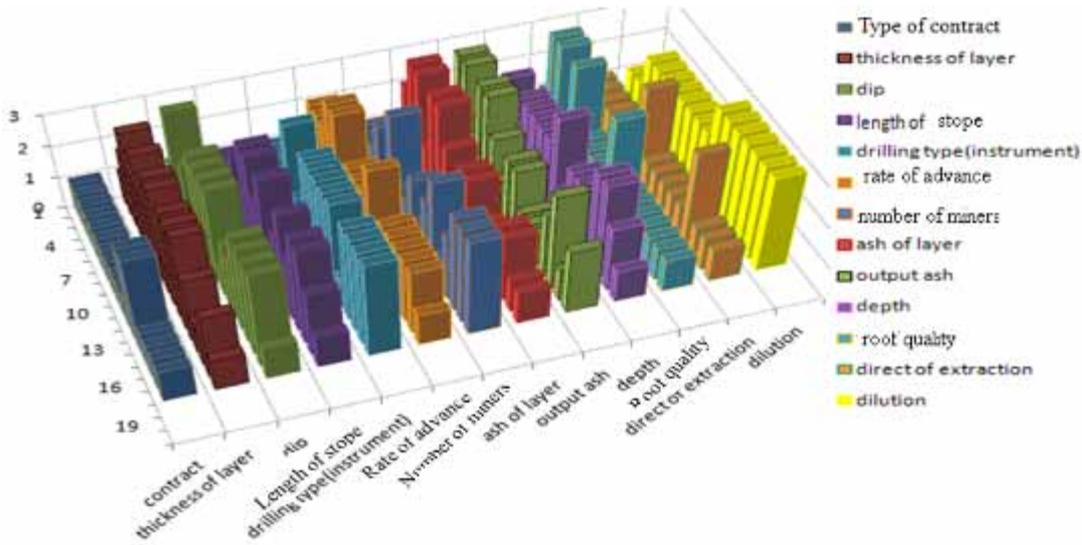

Figure 3. Result of transferring attributes in three categories over the training data set (vertical axis) by 1-D SOM

The obtained reduct set, by Johnson's reduct algorithm is: *{thickness of layer, length of stope, rate of advance, number of miners, type of extraction}*. Under this set, 19 rules are produced and on the test data are evaluated (figs4, 5). The quality and accuracy of rules can be identified using different criteria (φhm, 1998).

| Rule |
|---|
| thickness of layer(3) AND length of stope(1) AND rate of advance(2) AND number of miners(3) AND type of extraction(1) => dilution(1) |
| thickness of layer(3) AND length of stope(1) AND rate of advance(1) AND number of miners(1) AND type of extraction(1) => dilution(1) |
| thickness of layer(1) AND length of stope(2) AND rate of advance(2) AND number of miners(3) AND type of extraction(1) => dilution(1) |
| thickness of layer(1) AND length of stope(1) AND rate of advance(2) AND number of miners(1) AND type of extraction(1) => dilution(1) |
| thickness of layer(2) AND length of stope(1) AND rate of advance(2) AND number of miners(1) AND type of extraction(1) => dilution(1) |
| thickness of layer(3) AND length of stope(1) AND rate of advance(2) AND number of miners(1) AND type of extraction(1) => dilution(2) |
| thickness of layer(2) AND length of stope(2) AND rate of advance(3) AND number of miners(2) AND type of extraction(1) => dilution(2) |
| thickness of layer(3) AND length of stope(1) AND rate of advance(3) AND number of miners(1) AND type of extraction(1) => dilution(2) |
| thickness of layer(2) AND length of stope(3) AND rate of advance(1) AND number of miners(3) AND type of extraction(1) => dilution(2) |
| thickness of layer(3) AND length of stope(3) AND rate of advance(2) AND number of miners(1) AND type of extraction(1) => dilution(2) |
| thickness of layer(3) AND length of stope(2) AND rate of advance(2) AND number of miners(1) AND type of extraction(3) => dilution(2) |
| thickness of layer(1) AND length of stope(3) AND rate of advance(2) AND number of miners(2) AND type of extraction(1) => dilution(2) |
| thickness of layer(3) AND length of stope(2) AND rate of advance(1) AND number of miners(2) AND type of extraction(1) => dilution(3) |
| thickness of layer(3) AND length of stope(3) AND rate of advance(2) AND number of miners(3) AND type of extraction(1) => dilution(3) |
| thickness of layer(2) AND length of stope(2) AND rate of advance(2) AND number of miners(1) AND type of extraction(1) => dilution(3) |
| thickness of layer(1) AND length of stope(3) AND rate of advance(2) AND number of miners(2) AND type of extraction(3) => dilution(3) |
| thickness of layer(1) AND length of stope(3) AND rate of advance(2) AND number of miners(3) AND type of extraction(1) => dilution(3) |
| thickness of layer(2) AND length of stope(2) AND rate of advance(2) AND number of miners(3) AND type of extraction(1) => dilution(3) |
| thickness of layer(1) AND length of stope(1) AND rate of advance(1) AND number of miners(3) AND type of extraction(1) => dilution(3) |

Figure 4. The elicited rules -by RST (notice: here type of ascribed extraction type code transfer to 1&3)

Analysis of second situation is started off by setting number of close-open iteration and maximum number of rules equal to 15 and 4 in SONFIS-R, respectively. The error measure criterion in SONFIS is mean square error (MSE):

$$MSE = \frac{\sum_{i=1}^{m}(d_i^{real} - d_i^{classified})^2}{m} \quad (4)$$

where *m* is the number of test data.

Figure 6(a&b) indicate the results of the aforesaid system (so, performance of selected SONFIS-R on the test data). In this case, we set the range of first granules (crisp clusters) between 5 and 20, as well as lower and upper floor. So, the number of leanings in second layer of SONFIS is supposed as a constant value, i.e., 20, for all inserted crisp granules. Add to this, we use Gaussian membership functions in fuzzy clustering. After 45 time steps

(15 for each rule), the minimum MSE is come out in 17 neurons in SOM and 2 rules in NFIS (figures 7, 8 and table 2).

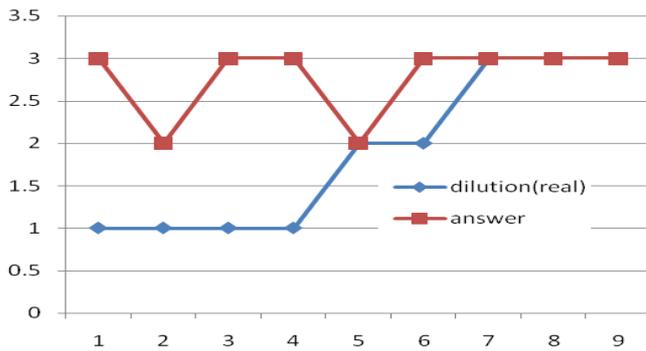

Figure 5. Answer of RST (based on the extracted rules) on the test data

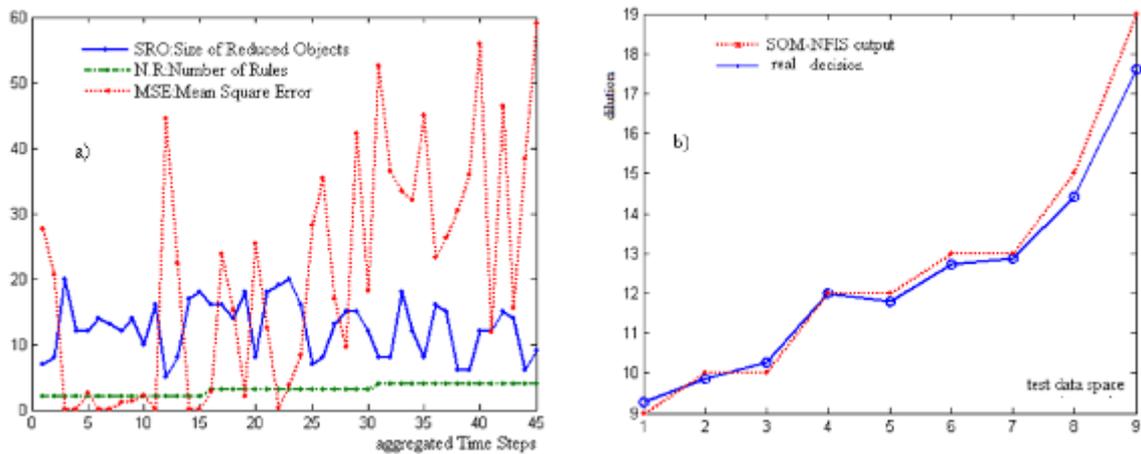

Figure 6. a) SONFIS-R results with maximum number of rules 4 and close-open iterations 10; b) Answer of selected (winner) SONFIS-R on the test data

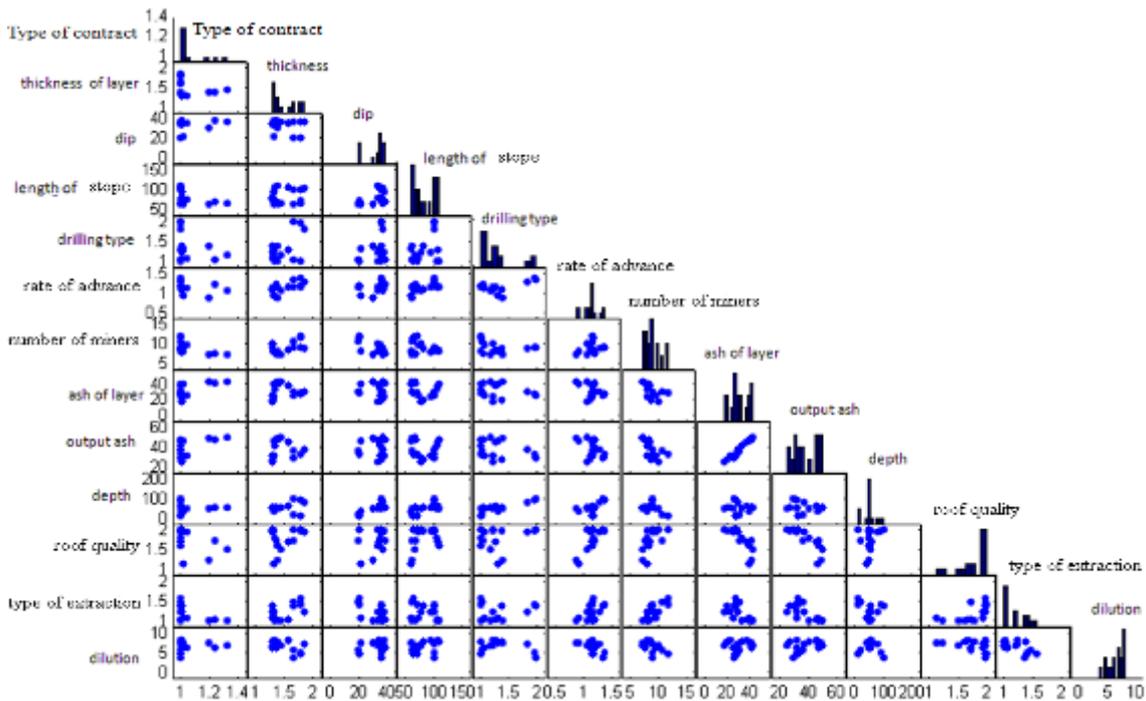

Figure 7. The results of clustered training data set by selected best SOM (17*1 neurons) by SONFIS-R

The results of first granulation by 17*1 neurons in competitive layer of SOM has been portrayed in figure 7, as matrix plot form. It must be notice here; we reduced all of objects in to the 17 patterns, which are in balance with the simplest rules of NFIS, while we had employed error measure criteria to balancing. SONFIS-R which has been employed in other comprehensive data set, show ability of this system in detection of the dominant structures on the attributes and representation of the simplest rules, as well as one wishes to catch up (Owladeghaffari et al,2008).

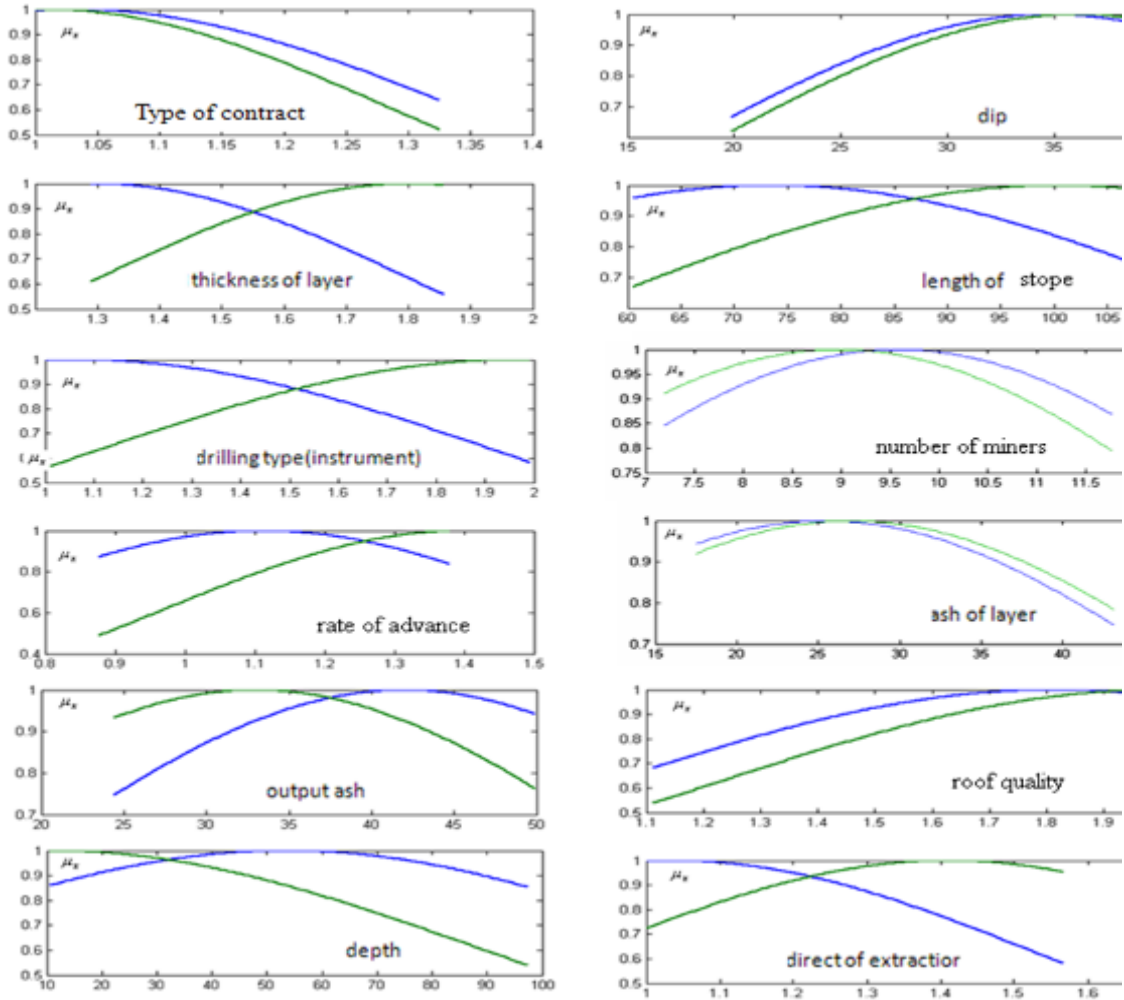

Figure 8. Final membership functions of inputs in SONFIS-R

Table2.Extrated rules by NFIS in SONFIS-R (*in i (i=1:12)* is agree with the decision table attributes and *mf j(j=1,2)* is proper with blue and green curves, respectively.

| |
|---|
| 1. If (in1 is in1mf1) and (in2 is in2mf1) and (in3 is in3mf1) and (in4 is in4mf1) and (in5 is in5mf1) and (in6 is in6mf1) and (in7 is in7mf1) and (in8 is in8mf1) and (in9 is in9mf1) and (in10 is in10mf1) and (in11 is in11mf1) and (in12 is in12mf1) then *(f1)* |
| 2. If (in1 is in1mf2) and (in2 is in2mf2) and (in3 is in3mf2) and (in4 is in4mf2) and (in5 is in5mf2) and (in6 is in6mf2) and (in7 is in7mf2) and (in8 is in8mf2) and (in9 is in9mf2) and (in10 is in10mf2) and (in11 is in11mf2) and (in12 is in12mf2) then *(f2)* |

It is noteworthy that the density of membership functions (fig8) and the histogram of variables appeared by SOM (fig7) have been coincided. Add to this in second granulation level
-Fuzzy rules- decision part of rules are in the linear (one order) form of premises (*f1&f2* in table2). In a supplementary way to take visual relates on the most effective parameters based on the reduct members, we use the fuzzy rules on those attributes and decision part (fig 9). Figure 9 shows how dilution changes under the main effective parameters, produced by RST as a reduct set, for instance increasing of length of stope and thickness of layer raises the dilution values as well as rate of advance and dip parameters. In other sense, transferring of advance face to backward will display a remarkable reduces in dilution rate.

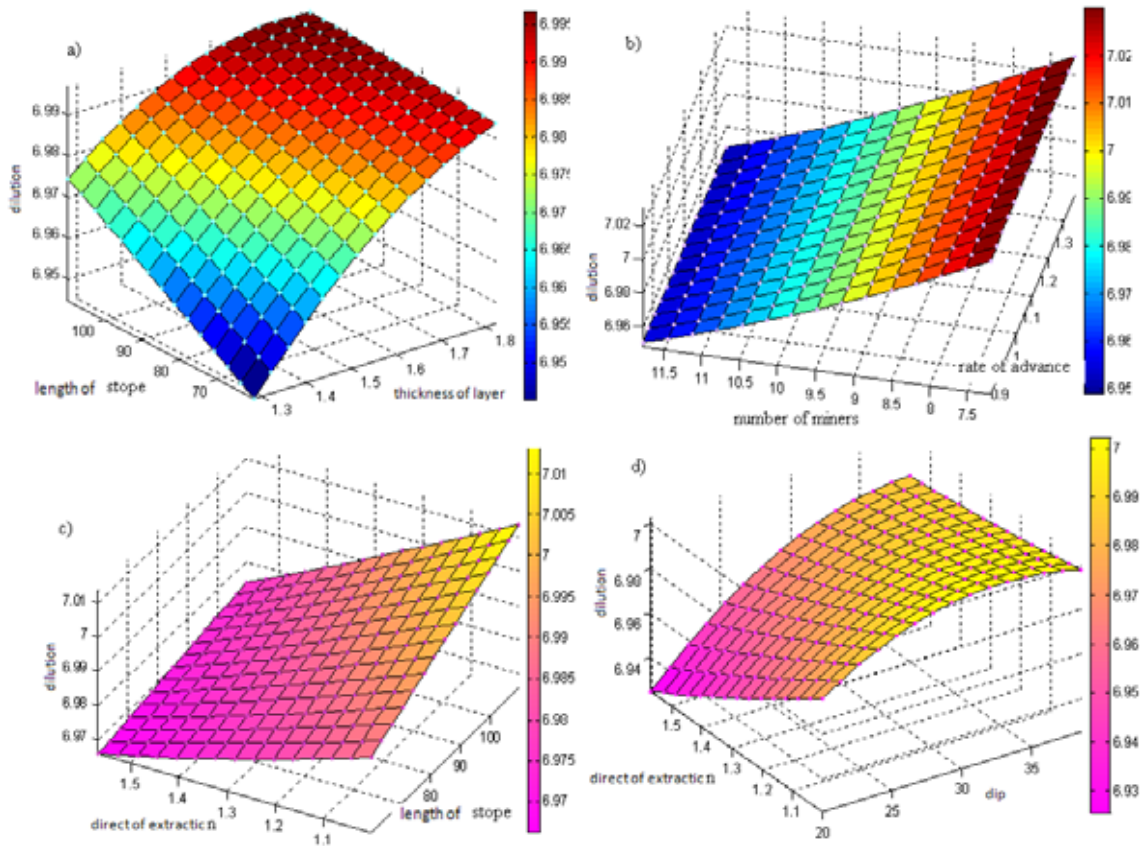

Figure 9.The 3D relations on the reduct set members and dilution by extracted rules of SONFIS-R

4 CONCLUSION

To better controlling of dilution rate in coal mines, considering of the high scope of variables will be undeniable. Under this view and "Information bears on uncertainty", we employed two main algorithms: Self Organizing Neuro-Fuzzy Inference System (Random) SONFIS-R and Rough Set Theory, to analysis of dilution in a few Iranian coal mines.

So, the reduced set by RST, among 13 effective parameters on decision part, has been obtained. These attributes upon the best fuzzy rules-in balance with reduced objects- exhibit more details of dependency variations, as if nearly concured with the crisp granules (SOM weights).The emerged results by utilizing mentioned methods shows that the high sensitive variables are "thickness of layer, length of stope, rate of advance, number of miners, type of extraction".